\documentclass[letterpaper, 10 pt, journal, twoside]{IEEEtran}
\usepackage{amsmath,amsfonts}
\usepackage{algorithmic}
\usepackage{algorithm}
\usepackage{array}
\usepackage{textcomp}
\usepackage{stfloats}
\usepackage{url}
\usepackage{verbatim}
\usepackage{graphicx}
\usepackage{cite}
\usepackage{graphicx,subfigure}
\usepackage{amssymb, gensymb}
\usepackage{hyperref}
\hypersetup{
    colorlinks=true,
    urlcolor=magenta,
}

\begin{document}

\title{Lifelong Topological Visual Navigation}


% \author{Rey Reza Wiyatno$^{\dagger, \ast}$, Anqi Xu$^{\ddagger}$, and Liam Paull$^{\dagger}$% <-this % stops a space
\author{Rey Reza Wiyatno$^{1}$, Anqi Xu$^{2}$, and Liam Paull$^{1}$
\thanks{Manuscript received: February, 24, 2022; Revised May, 21, 2022; Accepted June, 17, 2022.}%Use only for final RAL version
\thanks{This paper was recommended for publication by Editor Eric Marchand upon evaluation of the Associate Editor and Reviewers' comments.
% This work was supported by (organizations/grants which supported the work.)
}%Use only for final RAL version
\thanks{$^{1}$Rey Reza Wiyatno and Liam Paull are with Montr\'eal Robotics and Embodied AI Lab (REAL) and DIRO at the University of Montr\'eal, QC H3T 1J4, Canada, and Mila, QC H2S 3H1, Canada 
        {\tt\footnotesize rey.wiyatno@umontreal.ca, paulll@iro.umontreal.ca}}%
\thanks{$^{2}$Anqi Xu conducted this work with support from his past affiliation with Element AI, H2S 3G9, Canada
        {\tt\footnotesize anqi.xu@mail.mcgill.ca}}
% \thanks{$^{\ast}$ corresponding author}%
\thanks{Digital Object Identifier (DOI): see top of this page.}
}

\markboth{IEEE ROBOTICS AND AUTOMATION LETTERS. PREPRINT VERSION. ACCEPTED JUNE, 2022}%
{Wiyatno \MakeLowercase{\textit{et al.}}: Lifelong Topological Visual Navigation}


\thispagestyle{empty}
\onecolumn

This paper has been accepted to IEEE Robotics and Automation Letters (RA-L) and International Conference on Intelligent Robots and Systems (IROS) 2022.

\

© 2022 IEEE. Personal use of this material is permitted. Permission from IEEE must be obtained for all other uses, in any current or future media, including reprinting/republishing this material for advertising or promotional purposes, creating new collective works, for resale or redistribution to servers or lists, or reuse of any copyrighted component of this work in other works.
\setcounter{page}{0}

\twocolumn

\maketitle

\begin{abstract}
Commonly, learning-based topological navigation approaches produce a local policy while preserving some loose connectivity of the space through a topological map. Nevertheless, spurious or missing edges in the topological graph often lead to navigation failure. In this work, we propose a sampling-based graph building method, which results in sparser graphs yet with higher navigation performance compared to baseline methods. We also propose graph maintenance strategies that eliminate spurious edges and expand the graph as needed, which improves lifelong navigation performance. Unlike controllers that learn from fixed training environments, we show that our model can be fine-tuned using only a small number of collected trajectory images from a real-world environment where the agent is deployed. We demonstrate successful navigation after fine-tuning on real-world environments, and notably show significant navigation improvements over time by applying our lifelong graph maintenance strategies.\footnote{Project page: 
\href{https://montrealrobotics.ca/ltvn/}{https://montrealrobotics.ca/ltvn/}}
\end{abstract}


% \section{Introduction}
% \IEEEPARstart{T}{his} file is intended to serve as a ``sample article file''
% for IEEE journal papers produced under \LaTeX\ using
% IEEEtran.cls version 1.8b and later. The most common elements are covered in the simplified and updated instructions in ``New\_IEEEtran\_how-to.pdf''. For less common elements you can refer back to the original ``IEEEtran\_HOWTO.pdf''. It is assumed that the reader has a basic working knowledge of \LaTeX.

%%------------------------------------------------- %
\section{INTRODUCTION} \label{sec:intro}
%%------------------------------------------------- %

\IEEEPARstart{A}{} standard workflow for robot navigation involves first manually piloting a robot to build a metric map with simultaneous localization and mapping (SLAM)~\cite{leonard1991slam}. However, with this type of metric-based navigation, it is unintuitive to specify goals in metric space, and also tedious for an expert user to pilot the robot around to build the map. Ideally, navigation goals should have an intuitive representation, such as images of target objects or locations, and a non-expert user should be able to provide them in a natural way. While we see the emergence of learning-based methods that directly map images to actions by learning a global controller~\cite{zhu2017target}, these policies tend to be reactive, are not data efficient, and are not suitable for long-distance navigation. 

An alternative strategy is to forego the metric map and maintain a \emph{topological} representation of the environment~\cite{Kuipers_Levitt_1988}. In such a setup, each edge in the graph encodes the traversability between two locations, while a local controller is used to actually navigate the edge. In contrast to a global controller, navigating within a local vicinity is an easier task than navigating globally through a complex environment. The challenge here is how to construct such a representation in an efficient way that enables a local controller to navigate the environment.

% \IEEEpubidadjcol

A desirable setup is for the nodes in the topological graph to correspond directly to sensor data collected from the corresponding pose in space. We use colored depth (RGB-D) images as sensor data, and develop a model that jointly predicts reachability and relative transformation between two RGB-D images, which we will use to determine connectivity between the nodes of the graph. Importantly, we show that this model can be pre-trained using automatically generated simulated data, and then fine-tuned using only the data that is collected to build the graph in the target environment.

\begin{figure}[t]
    \centering
    \subfigure[Failed planning before graph maintenance]{\label{fig:sample_before}\includegraphics[width=0.42\textwidth]{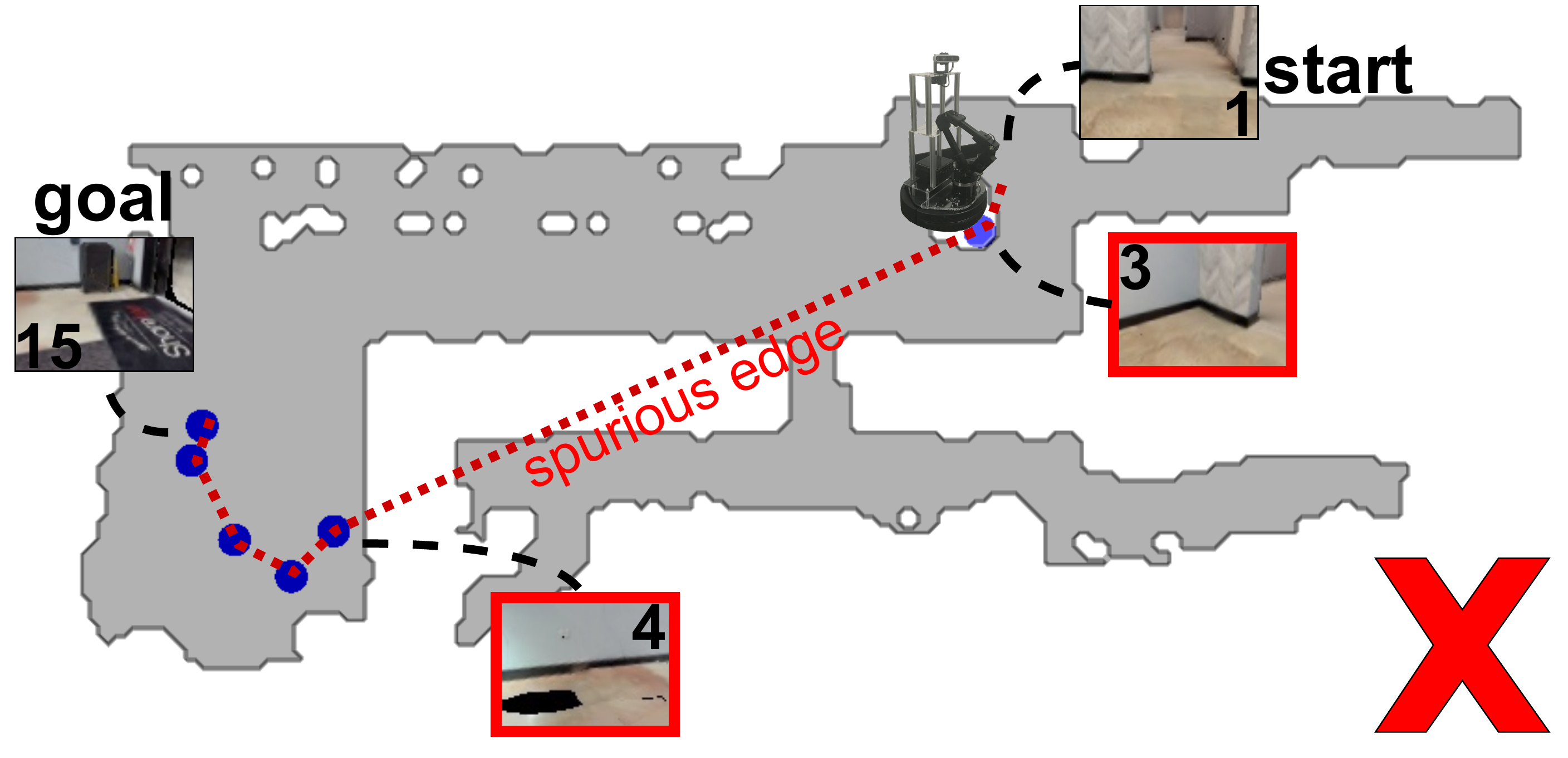}} % \hspace{2cm}
    \subfigure[Successful planning after graph maintenance]{\label{fig:sample_after}\includegraphics[width=0.42\textwidth]{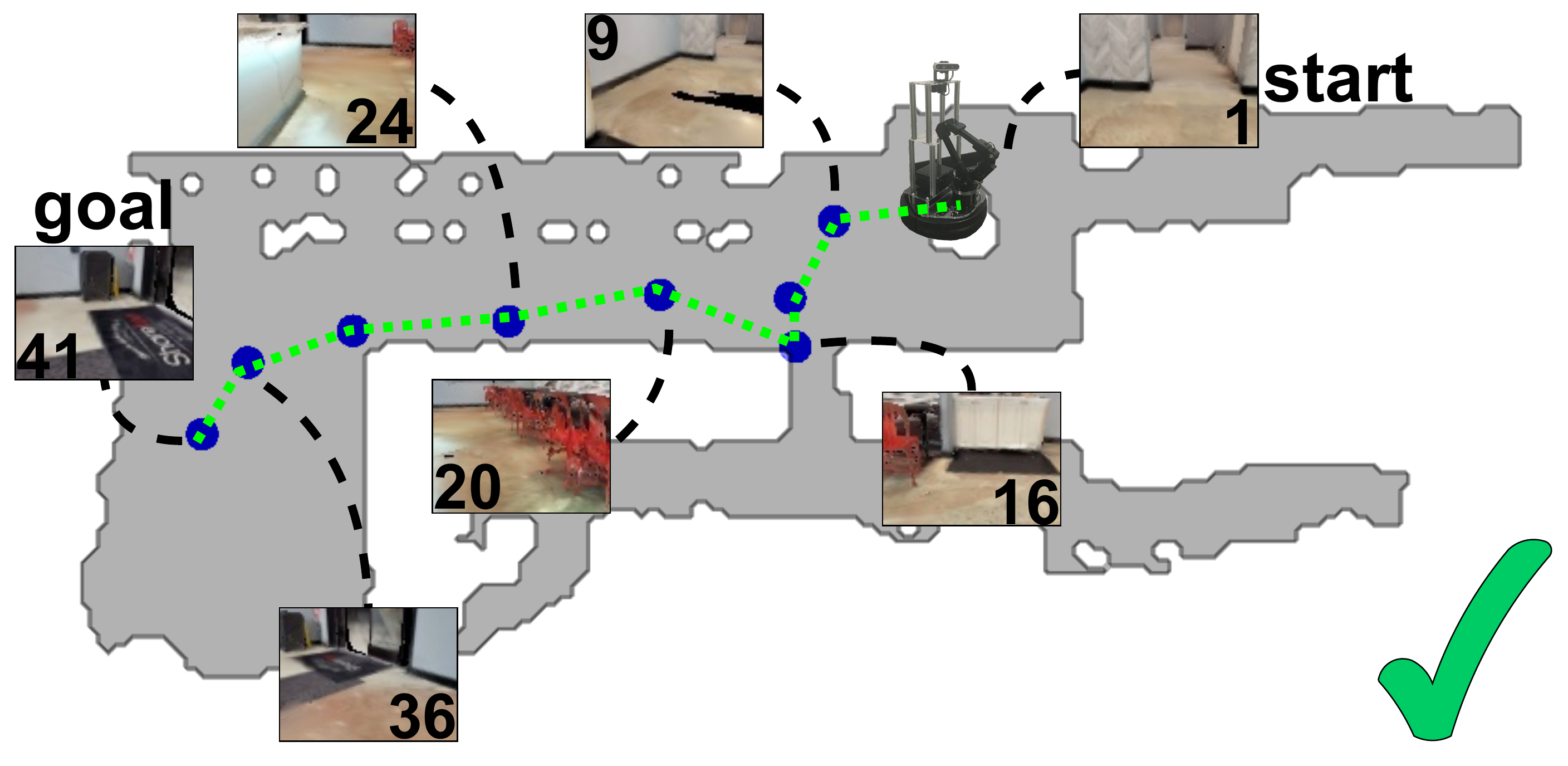}}
    \caption{\label{fig:sample_plan} Sample plans produced with our method to navigate from a start to a goal image, before and after graph maintenance. The blue dots indicate nodes within the planned path. The plan in Fig.~\ref{fig:sample_before} led to navigation failure since nodes \#3 and \#4 are erroneously connected. Fig.~\ref{fig:sample_after} showcases the refined plan after graph maintenance, which led to successful navigation.}
\end{figure}

To build the graph, we take inspiration from traditional sampling-based robotics planners such as probabilistic roadmaps (PRM)~\cite{kavraki1996prm} and rapidly-exploring random trees (RRT)~\cite{lavalle1998rapidly} but formulate the problem over \textit{sensor data} space rather than \textit{configuration} space.
We propose a sampling-based graph building process that produces sparser graphs and also improves navigation performance when compared to baselines. We construct this graph by sampling nodes from a pool of collected images and using the proposed model to determine the connectivity between the nodes.

Since the connectivity of our graph is determined by a potentially imperfect model, it is important to address the possibility of spurious edges. False positives from this model will induce spurious edges in the graph and may cause the agent to execute infeasible plans, while false negatives will result in edges being omitted and may result in failure to find a path when one actually exists. Thus, while other methods~\cite{savinov2018semiparametric, meng2020scaling, shah2021ving} treat their graphs as static objects, we continually refine ours based on what our agent experiences when executing navigation queries. As a result, these graph updates enable lifelong navigation; they eliminate spurious edges and possibly add new nodes that might be missing, as shown in Fig.~\ref{fig:sample_plan}, which improves navigation performance over time.

To summarize, our main contributions are: 

\begin{enumerate}
    \item A sampling-based graph building process that produces sparser graphs and improves navigation performance,
    \item A multi-purpose model for graph building, path execution, and graph maintenance, that can be fine-tuned in real-world using small amount of data,
    \item A graph maintenance procedure that enables continuous graph refinement during operation that improves lifelong navigation performance.
    % \item A graph maintenance procedure that improves lifelong navigation performance.
\end{enumerate}

%%------------------------------------------------- %
\section{RELATED WORKS} \label{sec:related}
%%------------------------------------------------- %

Learning-based approaches have shown promising results in solving visual navigation tasks. For example, several works have used reinforcement learning (RL) to learn to navigate based on a goal image~\cite{zhu2017target, 9370169}. Training RL policies requires significant computation and time however, and typically involve additional sim-to-real transfer method such as domain randomization~\cite{tobin2017domain} that in practice tend to not scale well in real-world. End-to-end methods also tend to not work well in long-distance navigation tasks.

% SPTM + ViNG
More closely related to our approach are semi-parametric topological memory (SPTM)~\cite{savinov2018semiparametric} and visual navigation with goals (ViNG)~\cite{shah2021ving}. SPTM builds a graph using a classifier that infers if two images are temporally close. However, the graph edges are unweighted, so false positive edges may be repeatedly chosen during planning. ViNG regresses the number of steps required to move from one image to another, and uses this to weigh each edge. ViNG also proposes to prune edges that are deemed by their model to be easily traversable during the graph building stage. In contrast, our pruning strategy operates \textit{continually} based on what our agent experiences when executing a navigation query, which leads into lifelong navigation improvements. Furthermore, while ViNG demonstrates the ability to navigate in the real-world, ViNG requires 40 hours of offline real-world data, which is tedious to gather. Our model can be fine-tuned in real-world using a significantly smaller dataset.

As a common concern, both SPTM and ViNG build a graph using \emph{all} images within the collected trajectories, which poses scalability and false connectivity issues. Furthermore, both methods build the graph without considering the capability of their controller, which may result in edges that are not traversable in practice. Moreover, by solely relying on temporal distance within collected trajectories, they are blind to the connection of nodes that are spatially close, yet temporally far within the explored trajectories (i.e., loops). In contrast, our graph building process relies on a model that is aware of the limitations of the controller used.

% BRM
Bayesian Relational Memory (BRM)~\cite{wu2019bayesian} builds a fully-connected graph where nodes and edges map to room types and the probability of room connectivity. BRM trains a classifier that predicts the probability of an image belonging to different room types. As the agent navigates, edge weights are refined using Bayesian updates. Our graph maintenance strategy is similar to that of BRM, but we can also introduce new nodes to the graph as necessary to enable planning.

% Meng et al
Meng \emph{et al.}~\cite{meng2020scaling} proposed a controller-dependent graph building method. At its core, a classifier is trained based on the controller rollout outcome in simulation to predict if an image pair is reachable. To build the graph, this classifier model is used to first sparsify highly reachable redundant nodes in the trajectories. Then, the remaining nodes are connected with edges weighted by predicted reachability scores. As a drawback, it is impractical to fine-tune this reachability model in the real-world, as it would require empirically unsafe controller rollouts between location pairs.

% Other value-based graphs
Other methods rely on an actor-critic model to evaluate graph connectivity using the critic~\cite{nasiriany2019planning, eysenbach2019search, scott2020sparse}. Scott \emph{et al.}~\cite{scott2020sparse} further sparsify the graph by only adding perceptually distinct nodes, merging nodes with shared connections, limiting the number of edges per node, and removing edges predicted as not traversable during test time. However, these sparsification strategies may lead to excessive false negative edges and poor connectivity. Also, such simulation-trained policies may not transfer well to real-world environments.

\begin{figure}[t]
    \centering
    \includegraphics[width=0.48\textwidth]{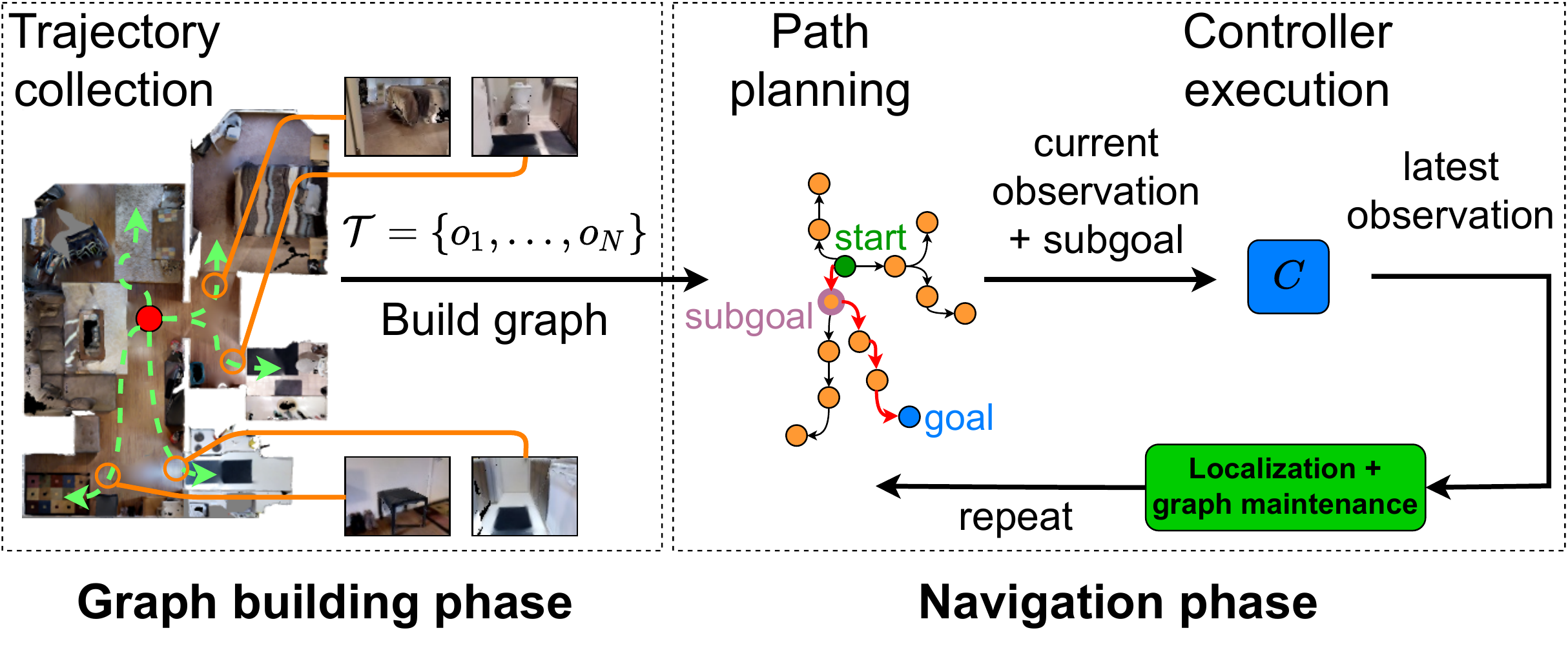}
    \caption{\label{fig:framework}A common topological navigation framework consists of separate graph building and navigation phases. When deployed in a new environment, an agent first collects observations from an environment and builds a topological graph. During navigation, the agent localizes itself on the graph, plans a path to a given goal, and moves to a subgoal using a controller. The agent then relocalizes itself, and repeats the planning and control steps until it reaches the goal. Our work highlights the importance of updating the graph, which is missing from most existing work.
    }
\end{figure}

Our system shares a common structure with other learning-based topological navigation methods, as shown in Fig.~\ref{fig:framework}. Still, our approach differs in choices for the learned model, data collection procedure, graph building approach, what graph edges encode, the controller used, and the graph maintenance strategy. We summarize these differences in Table~\ref{tab:compare_methods}. We shall show that our design choices lead to superior performance when deployed in various simulated environments, and also demonstrate strong system performance in the real-world.

\begin{table*}[t]
    \scriptsize
    \centering
    \caption{Comparison of various learning-based topological navigation methods for image-goal navigation tasks.}
    \begin{tabular}{|p{0.09\textwidth}|p{0.10\textwidth}|p{0.125\textwidth}|p{0.125\textwidth}|p{0.11\textwidth}|p{0.151\textwidth}|p{0.11\textwidth}|}
        \hline          &      &  &    &  &     &       \\
         [-2.5mm]
         & \textbf{Controller} & \textbf{Node selection} & \textbf{Edge weight} & \textbf{Path planner} & \textbf{Graph maintenance} & \textbf{Model fine-tuning} \\
        \hline          &      &  &      & &   &          \\ [-2.5mm]
        \textbf{SPTM}~\cite{savinov2018semiparametric} & Inverse dynamics & All nodes & Temporal, unweighted & Graph search & None & Self-supervised \\
        \hline          &      &  &      & &    &         \\
        [-2.5mm]
        \textbf{HTM}~\cite{pmlr-v119-liu20h} & Inverse dynamics & All nodes & Contrastive loss & Graph search & None & None \\
        \hline          &      &  &      & &     &         \\
        [-2.5mm]
        \textbf{Meng \emph{et al.}}~\cite{meng2020scaling} & Potential-based & Incrementally selected & Reachability score & Graph search & None & None \\
        \hline          &      &  &      & &    &        \\
        [-2.5mm]
        \textbf{LEAP}~\cite{nasiriany2019planning} & RL & Optimization-based & Value function & Optimization-based & None & None \\
        \hline          &      &  &      & &    &          \\
        [-2.5mm]
        \textbf{SoRB}~\cite{eysenbach2019search} & RL & All nodes & Value function & Graph search & None & None \\
        \hline          &      &  &      & &    &        \\
        [-2.5mm]
        \textbf{SGM}~\cite{scott2020sparse} & RL & Incrementally selected & Value function & Graph search & Edge pruning & None \\
        \hline          &      &  &      & &    &        \\
        [-2.5mm]
        \textbf{ViNG}~\cite{shah2021ving} & Position-based & All nodes & Temporal & Graph search & None & Self-supervised \\
        \hline          &      &  &      & &    &        \\
        [-2.5mm]
        \textbf{Ours} & Position-based & Sampling-based & Pose-based distance & Graph search & Edge update, node addition & Self-supervised \\
        \hline
    \end{tabular}
    \label{tab:compare_methods}
\end{table*}

%%------------------------------------------------- %
\section{PROPOSED METHOD} \label{sec:method}
%%------------------------------------------------- %

Our work focuses on navigation tasks where the goal is specified by a target RGB-D image. Following the framework in Fig.~\ref{fig:framework}, during graph building, we first execute a trajectory collection phase to obtain RGB-D images $\mathcal{T} = \{o_1, ..., o_N\}$. We then use $\mathcal{T}$ to build a graph $G=(V,E)$, where vertices $V$ are images and directed edges $E$ represent traversability.

During navigation, we present the agent with a goal image $o_{g}$. The agent first localizes itself on the graph based on its current observation $o_{a}$, plans a path to $o_{g}$, picks a subgoal observation $o_{sg}$, and moves towards it using its controller. The agent then relocalizes itself on the graph using its latest observation, and updates the graph based on its experience. These processes are repeated until the agent reaches $o_g$.

\begin{figure*}[t]
    \centering
    \subfigure[Our model]{\label{fig:our_model}\includegraphics[width=0.225\textwidth]{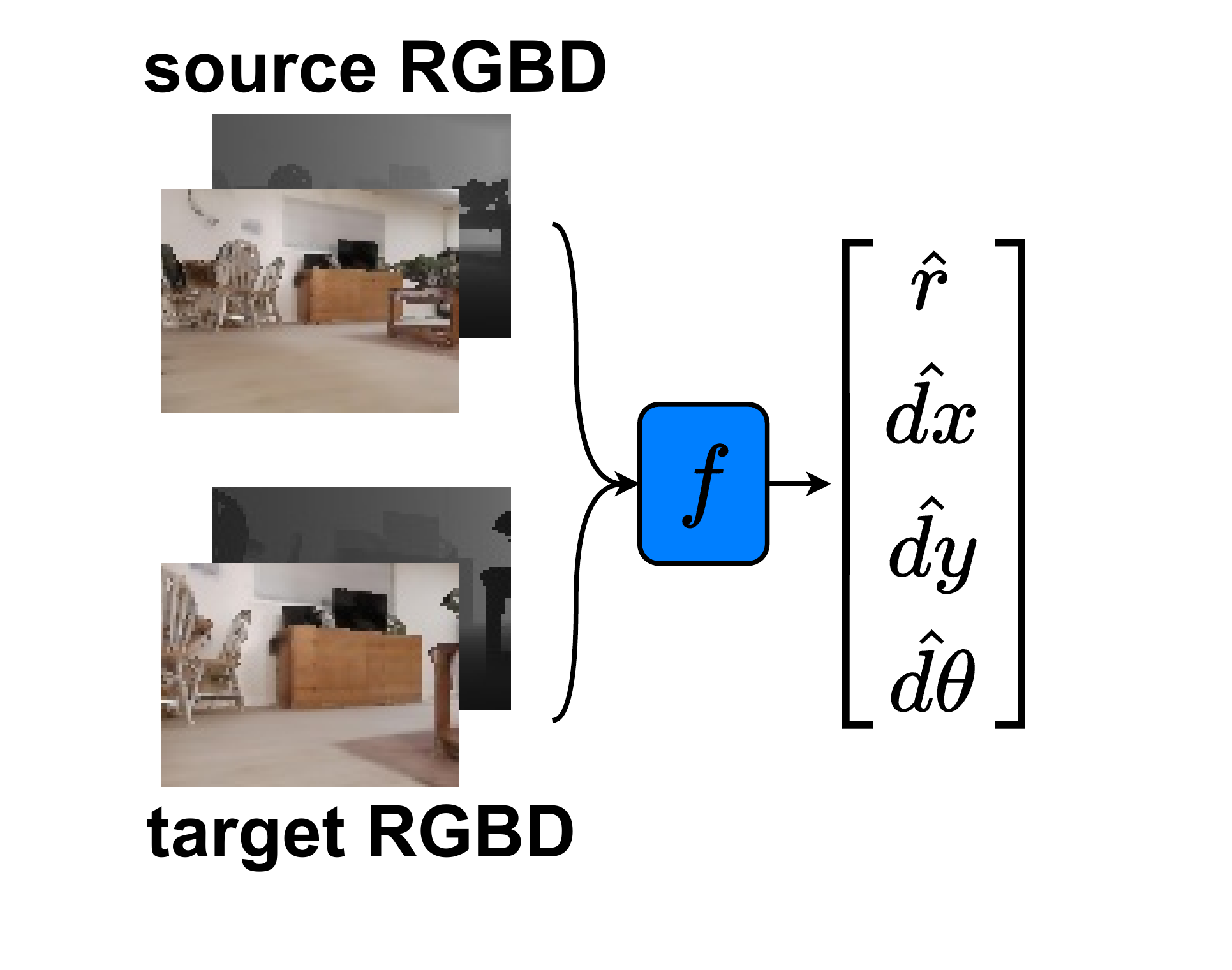}}
    \subfigure[Automated dataset creation]{\label{fig:dataset}\includegraphics[width=0.5\textwidth]{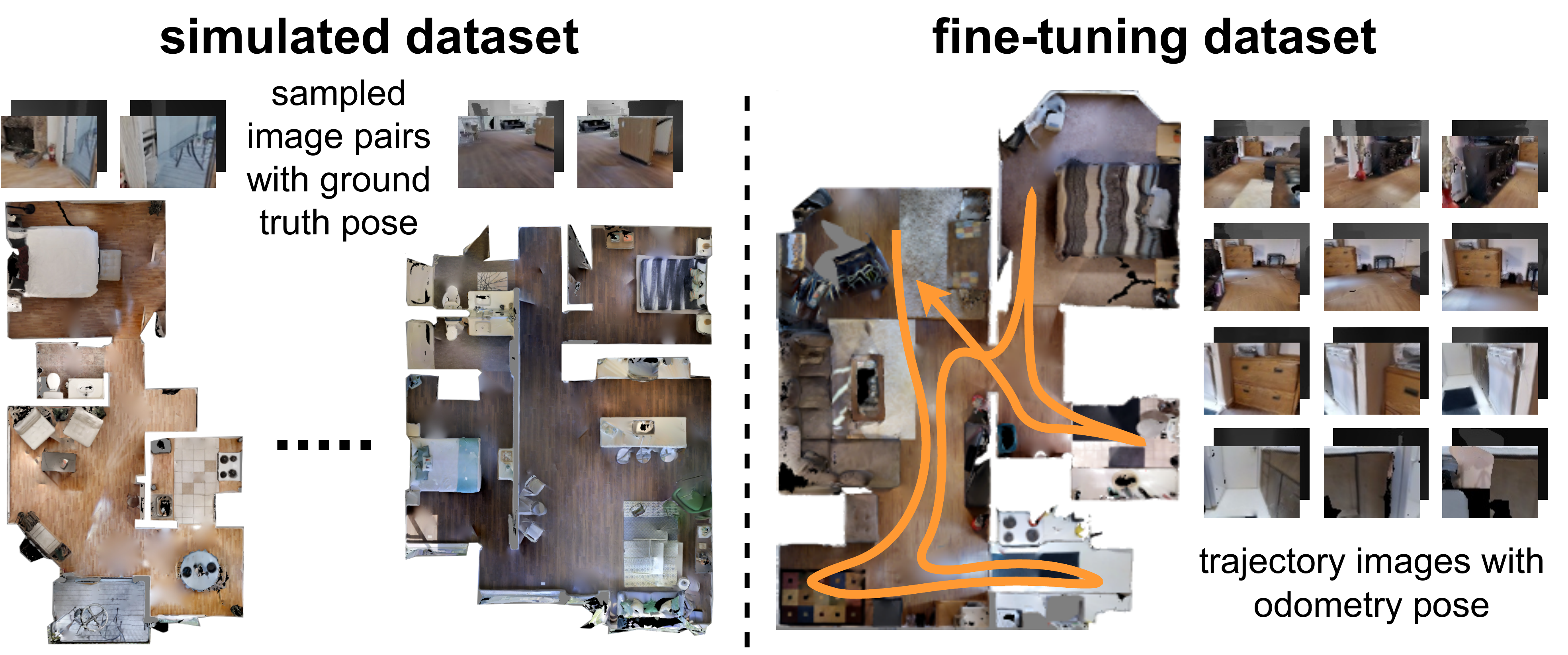}}
    \subfigure[Sampling-based graph building]{\label{fig:graph_building}\includegraphics[width=0.225\textwidth]{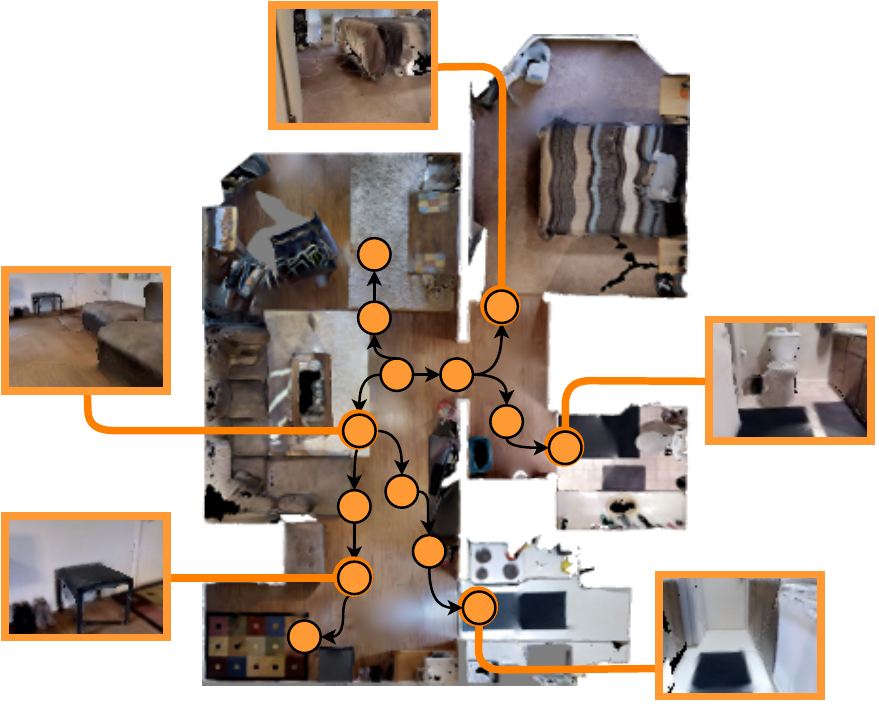}}
    \subfigure[Connectivity and edge update]{\label{fig:graph_update_1}\includegraphics[width=0.35\textwidth]{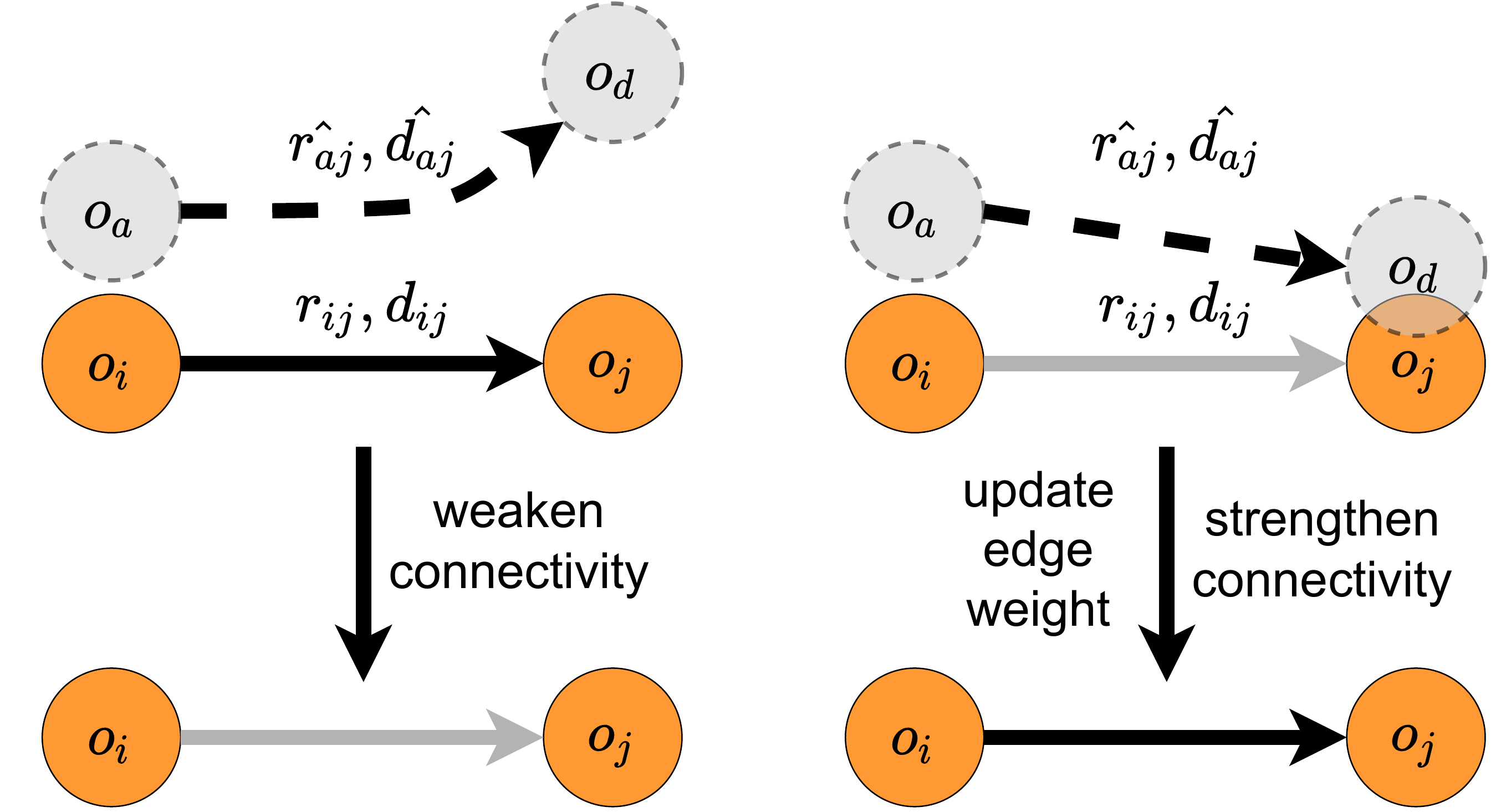}}
    \subfigure[Sampling-based graph expansion]{\label{fig:graph_update_2}\includegraphics[width=0.35\textwidth]{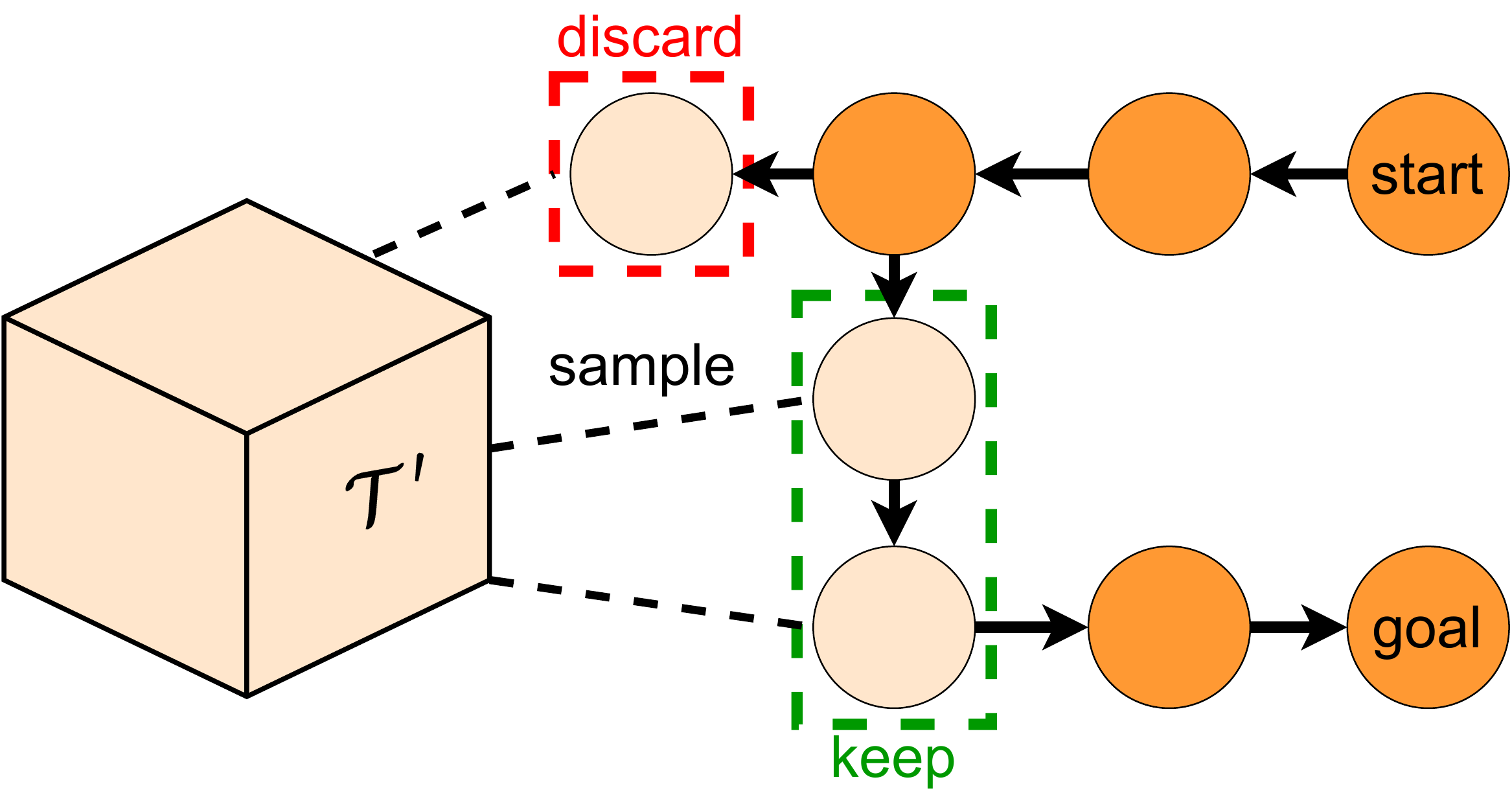}}
    \caption{\label{fig:proposed_method}
    Illustrations of our main contributions. Fig.~\ref{fig:our_model} depicts the proposed model, which takes source and target RGB-D images and outputs a reachability score $\hat{r}$ and a waypoint $\hat{w} = [\hat{dx}, \hat{dy}, \hat{d\theta}]$. 
    As shown in Fig.~\ref{fig:dataset}, we first train our model in simulation by collecting RGB-D image pairs from different environments. Later, we fine-tune our model in a new environment using only RGB-D images and their corresponding odometry data from a collected trajectory. 
    Our sampling-based graph building method is depicted in Fig.~\ref{fig:graph_building}, where we only select a portion of trajectory images to build the graph.
    Fig.~\ref{fig:graph_update_1} illustrates one of the proposed graph maintenance methods that updates the graph based on the success of an agent in traversing an edge. If the agent fails, the edge connectivity is weakened, else, its connectivity is strengthened and its weight is updated. Fig.~\ref{fig:graph_update_2} illustrates the second graph maintenance method that expands the graph by sampling from the remaining trajectory data $\mathcal{T'}$ to enable planning when the agent is unable to find a path.}
\end{figure*}

The rest of this section discusses our main contributions, which are illustrated in Fig.~\ref{fig:proposed_method}. First, we present a simple yet versatile model that is the crux of our navigation system, be it for graph building, path execution, or graph maintenance. We then discuss our proposed sampling-based graph building algorithm that produces sparser graphs compared to baselines, and how to perform navigation with the proposed model. Finally, we present lifelong graph maintenance strategies that lead to improved navigation performance as our agent executes more queries in a target environment.

%-----------------------------------------%
\subsection{Reachability and Waypoint Prediction Model} \label{sec:model}
%-----------------------------------------%

Our goal is to design a model that we can use in most of the navigation aspects. We train a convolutional neural network $f(o_i, o_j) = [\hat{r}, \hat{dx}, \hat{dy}, \hat{d\theta}]$ that takes two RGB-D images $(o_i, o_j)$ and jointly estimates reachability $r \in \{0,1\}$ from one image to another, and their relative transformation represented as a waypoint $w = [dx, dy, d\theta] \in \mathbb{R}^{2} \times (-\pi, \pi]$. To simplify the pose estimation problem, the waypoint only contributes to the training loss for reachable data points.

This model is used in a number of the components of our system. First, we use our model for graph building by using the reachability and pose estimates to determine the node connectivity and edge weights, respectively. We also use our model to perform localization and graph maintenance. Furthermore, we use a position-based feedback controller to navigate to waypoints predicted by our model.

We train our model with full supervision in simulation on a broad range of simulated scenes. Additionally, we can later fine-tune our model using only the trajectory data acquired from the environment where the agent is deployed. As a result, we can use our model in the real-world environment without needing to tediously collect and manually label a large amount of real-world data. We discuss how we create both simulated and fine-tuning datasets in Section~\ref{sec:dataset}.

We train the proposed model by minimizing the binary cross-entropy for reachability and regression loss for the relative waypoint. Concretely, the loss functions are 

\begin{equation}
\begin{gathered} \label{eq:training_loss}
L_{r}(r, \hat{r}) = -(r \log (\hat{r}) + (1 - r) \log (1 - \hat{r})),
\\
L_{p}(dx, dy, \hat{dx}, \hat{dy}) = ||[dx, dy] - [\hat{dx},\hat{dy}]||_2,
\\
L_{\theta}(d\theta, \hat{d\theta}) =|\sin(d\theta) - \sin(\hat{d\theta})| + |\cos(d\theta) - \cos(\hat{d\theta})|,
\\
L_{total}(r, dx, dy, d\theta, \hat{r}, \hat{dx}, \hat{dy}, \hat{d\theta}) = L_{r}(r, \hat{r}) \\ + r \big(\alpha L_{p}(dx, dy, \hat{dx}, \hat{dy}) + \beta L_{\theta}(d\theta, \hat{d\theta}) \big),
\end{gathered}
\end{equation}

\noindent where $L_{r}(r, \hat{r})$ is the reachability loss, $L_{p}(dx, dy, \hat{dx}, \hat{dy})$ is the position loss, and $L_{\theta}(d\theta, \hat{d\theta})$ is the rotation loss. Variables $r$, $dx$, $dy$, $d_{\theta}$ are the ground truth labels for the reachability and the relative waypoint predictions, whereas $\alpha$ and $\beta$ are hyperparameters to weigh the loss terms. 

%-----------------------------------------%
\subsection{Automated Dataset Creation} \label{sec:dataset}
%-----------------------------------------%

We aim to create a diverse dataset such that our model can generalize well to real-world environments without collecting a large real-world dataset. We thus create a dataset by sampling image pairs from various simulated environments.

In simulation, the waypoint label can be computed easily since the absolute pose of each observation is known. For reachability between two RGB-D observations, similar to Meng \emph{et al.}~\cite{meng2020scaling}, we define node-to-node reachability to be controller-dependent. Nevertheless, instead of rolling out a controller to determine reachability, we assume a simple control strategy based on motion primitives (i.e., Dubins curves), which allows us to compute reachability analytically. 

\begin{figure}[h]
    \centering
    \includegraphics[width=0.35\textwidth]{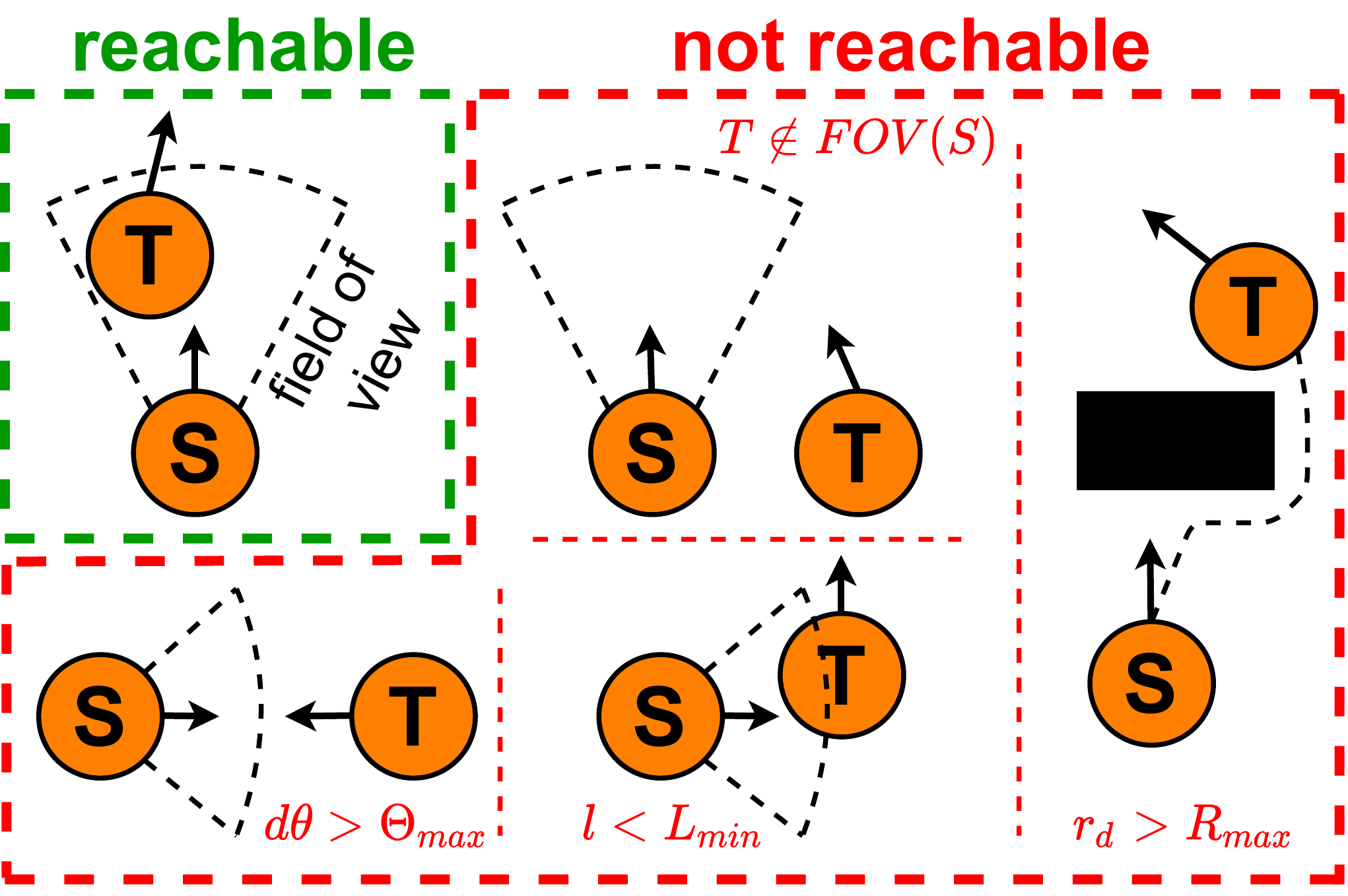}
    \caption{\label{fig:reachable} Illustration of reachable and non-reachable situations for a pair of source (\textbf{S}) and target (\textbf{T}) nodes.
    }
\end{figure}

We determine the reachability label based on visual overlap and spatial distance between two observations. Fig.~\ref{fig:reachable}, illustrates various reachable and non-reachable situations during data collection in simulation. Specifically, two observations are labeled as reachable if:
\begin{enumerate}
    \item The visual overlap ratio between the two images, $l$, is larger than $L_{min}$, computed based on depth data;
    \item The ratio of the shortest feasible path length over Euclidean distance between the poses, $r_{d}$, is smaller than $R_{max}$, to filter out obstacle-laden paths;
    \item The target pose must be visible in the initial image, so that our model can visually determine reachability;
    \item The Euclidean distance to the target must be less than $E_{max}$, and the relative yaw must be less than $\Theta_{max}$.
\end{enumerate}

\noindent During training, we define $o_j$ to be reachable only if it is in front of $o_i$. Yet, when navigating, the agent can move from $o_j$ to $o_i$ by following the predicted waypoint $w$ in reverse.

Because of how we define reachability, we can sample pairs of observations independently from various environments. Thus, our dataset creation in simulation follows the common independently and identically distributed assumption when training a machine learning model. This is in contrast to SPTM and ViNG where each datapoint is sampled from sequential image trajectories that are obtained from an agent operating in an environment following a random policy.

A key feature of our method is the ability to fine-tune the proposed model in any target domain, by using the same trajectory data $\mathcal{T}$ that we use to build the graph. Although SPTM and ViNG can also be trained on target-domain trajectories, our model does not need to be trained on a large real-world dataset, as it has already been trained within various simulated environments. In order for fine-tuning to be practical in the real-world, we only assume that the collected trajectories must have associated pose odometry to substitute for ground-truth pose data. Thankfully, odometry is readily available from commodity sensors such as inertial measurement units or wheel encoders.

Since visual overlap and shortest feasible path length are no longer accessible during fine-tuning, as a proxy criterion to determine reachability, we take an observation pair $(o_i, o_j) \in \mathcal{T}$ where $j > i$ and check if they are separated by at most $H$ time steps during trajectory collection. While the use of odometry as a supervisory signal for pose estimation can be noisy, the long-term pose drift should be minimal since reachable waypoints must be temporally close.

%-----------------------------------------%
\subsection{Sampling-based Graph Building} \label{sec:graph_building}
%-----------------------------------------%

A dense graph is inefficient to search over, and is also likely to exhibit spurious edges, which is a common cause of failure in topological navigation. Our goal is to build a graph with a minimum number of nodes and edges without sacrificing navigation performance. Thus, instead of using all images in $\mathcal{T}$, we build our graph incrementally via sampling.

Algorithm~\ref{alg:graph_building} describes the proposed graph building process. We initialize the graph as a randomly drawn node $o \in \mathcal{T}$. We also initialize a copy of the trajectory data $\mathcal{T'} = \mathcal{T} \setminus o$ to keep track of nodes that are not yet added to the graph. In each iteration, we sample a node $o_{r} \in \mathcal{T'}$ (or equivalently, sampling from shuffled $\mathcal{T'}$), check if it can be merged with or connected to existing graph vertices, and if so, remove it from $\mathcal{T'}$. If $o_{r}$ can be connected with any of the existing nodes $o_j \in V$, we weigh the edge with a distance function based on the relative pose transformation between the pair as predicted by the model $f$. Concretely, we define the distance to a waypoint $w$~\cite{barfoot2017book} as

\begin{equation} \label{eq:distance}
d(w) = ||\log T(w)||_{F},
\end{equation}

\noindent where $T(\cdot)$ converts a waypoint into its matrix representation in $SE(2)$\footnote{A matrix in the form of $\begin{bmatrix}
\mathbf{R} & \mathbf{t}\\
\mathbf{0}_{1 \times 2} & 1\\
\end{bmatrix}$, where $\mathbf{R} \in \mathbb{R}^{2 \times 2}$ denotes the rotation matrix, and $\mathbf{t} \in \mathbb{R}^{2}$ denotes the translation vector.}, and $||\cdot||_{F}$ computes the Frobenius norm. This procedure continues until no more nodes can be added. 

%----%
\begin{algorithm}[t]
   \caption{Graph Building}
   \label{alg:graph_building}
\begin{algorithmic}
   \STATE {\bfseries Input:} Trajectory data $\mathcal{T}$
   \STATE {\bfseries Init.:} $V = \{o \in \mathcal{T}\}$, $E = \varnothing$, $u = \texttt{True}$, $\mathcal{T'} = \mathcal{T} \setminus o$
   \WHILE{$u == \texttt{True}$}
   \STATE $u = \texttt{False}$
   \FOR{$o_r \in \texttt{shuffled}(\mathcal{T'})$}
   \IF{\texttt{isMergeable}$(o_r, V)$}
   \STATE $\mathcal{T'} = \mathcal{T'} \setminus o_{r}$
   \ELSE
   \STATE $c = \texttt{False}$
   \FOR{all $o_j \in V$}
   \IF{\texttt{isConnectable}$(o_r, o_j)$}
   \STATE $V, E = V \cup o_{r}, E \cup (o_{r}, o_j)$
   \STATE $c = \texttt{True}$
   \ENDIF
   \ENDFOR
   \IF{$c$}
   \STATE $u = \texttt{True}$
   \STATE $\mathcal{T'} = \mathcal{T'} \setminus o_{r}$
   \ENDIF
   \ENDIF
   \ENDFOR
   \ENDWHILE
   \STATE {\bfseries Return:} ($V$, $E$), $\mathcal{T'}$
\end{algorithmic}
\end{algorithm}
%----% 

To build a sparse yet reliable graph, we would like to connect nodes that are close together, but not the ones that are too close. To this end, we introduce two operators: $\texttt{isMergeable}$ and $\texttt{isConnectable}$. First, $\texttt{isMergeable}$ assesses whether a node is \emph{excessively} close to existing nodes and thus can be thrown away for being redundant. Second, $\texttt{isConnectable}$ checks if a node is \emph{sufficiently} close to another node such that a local controller can successfully execute the edge directly. These two distance thresholds are controlled by empirically-tuned hyperparameters $D_m$ and $D_c$. Due to the proposed node merging mechanism, our graph building method results in a sparser graph compared to other methods.

%-----------------------------------------%
\subsection{Navigation} \label{sec:navigation}
%-----------------------------------------%

Here, we describe how we can execute a navigation query with our model $f$ and the graph $G$. We first localize the agent on the graph based on its current observation $o_a$. Concretely, we use $f$ to compare pairwise distances between $o_a$ and all nodes in the graph, and identify the closest vertex where the distance is below $D_{\ell}$. To save computational cost, we first attempt to localize locally by considering only directly adjacent vertices from nodes within the last planned path, and then reverting to global localization if this fails. 

For planning, we use Dijkstra's algorithm~\cite{dijkstra1959} to find a path from where $o_a$ is localized to a given goal node $o_g \in V$, and select the first node in the path as subgoal $o_{sg}$. We then predict the waypoint from $o_{a}$ to $o_{sg}$, and use a position-based feedback controller to reach $o_{sg}$. At the end of controller execution, we take the agent's latest observation to relocalize the agent, and perform the proposed graph maintenance to refine the graph, as will be described in Section~\ref{sec:graph_update}. These are then repeated until the agent arrives at $o_g$.

%-----------------------------------------%
\subsection{Lifelong Graph Maintenance} \label{sec:graph_update}
%-----------------------------------------%

We propose two types of continuous graph refinements to aid navigation performance. The first is a method to correct graph edges based on the success of an agent in traversing an edge. This results in the removal of spurious edges and enhanced connectivity of traversable edges. Second, we add \emph{new} nodes to the graph either when observations are novel or when we cannot find a path to a goal.

We define two properties associated with each edge between physical locations $i$ and $j$ that are revised during graph maintenance: an edge's true connectivity after the $t$-th update modeled as $r^t_{ij} \sim \textrm{Bernoulli}(p^t_{ij})$, and an edge's distance weight modeled as $d^t_{ij} \sim \mathcal{N}(\mu^t_{ij}, (\sigma^t_{ij})^2)$. These are initialized respectively as $p^0_{ij} = \hat{r}^0_{ij}$, $\mu^0_{ij} = d(\hat{w}^0_{ij})$, and $(\sigma^0_{ij})^2 = \sigma^2$, where $(\hat{r}^0_{ij}, \hat{w}^0_{ij}) = f(o_i, o_j)$ are the predictions of our model during graph building, and $\sigma^2$ is derived empirically from the variance of our model's distance predictions across a validation dataset. We further define the probability of successful traversal through an edge as $p(s)$, where the conditional likelihood of the edge's existence $p(s|r)$ is also empirically determined.

Fig.~\ref{fig:graph_update_1} depicts how we update these edge properties after each traversal attempt. Given the agent's observation $o_a$ that is localized to $o_i$ on the graph, a target node $o_j$, the agent's latest observation after edge traversal $o_d$, we determine success of traversal via $\texttt{isConnectable}(o_d, o_j)$. We then update the edge's connectivity using discrete Bayes update:

\begin{equation} \label{eq:bayes}
p(r^{t+1}_{ij}|s) = \frac{p(s|r^t_{ij})p(r^t_{ij})}{p(s)}.
\end{equation}

\noindent When the agent fails to reach $o_j$, we prune the edge when $p(r^{t+1}_{ij}|s) < R_{p}$. Upon a successful traversal, we also use the predicted distance $d(\hat{w}_{aj})$ between $o_a$ and $o_j$ to compute the weight posterior with Gaussian filter:

\begin{equation}
\begin{gathered} 
\mu^{t+1}_{ij} = \frac{\sigma^2}{(\sigma^t_{ij})^2 + \sigma^2} \mu^t_{ij} + \frac{(\sigma^t_{ij})^2}{(\sigma^t_{ij})^2 + \sigma^2} d(\hat{w}_{aj}),\\
(\sigma^{t+1}_{ij})^2 = \Big( \frac{1}{(\sigma^t_{ij})^2} + \frac{1}{\sigma^2} \Big)^{-1}.
\end{gathered}\label{eq:bayes_gaussian}
\end{equation}

\noindent In this way we can correct for erroneous edges based on what the agent actually experiences during navigation.

To expand the graph, if an observation cannot be localized, we consider it as novel and add it to the graph. Separately, Fig.~\ref{fig:graph_update_2} depicts how we expand our graph when a path to a goal is not found during navigation. In this situation, we iteratively sample new nodes from the remaining trajectory data $\mathcal{T'}$ until a path is found, and store them into a candidate set $\tilde{V}$. Denoting the nodes within the path as $V_p$, we then add \textit{only} the new nodes that are along the found path $\tilde{V} \cap V_p$ to the graph permanently and remove them from $\mathcal{T'}$, while returning other nodes $\tilde{V} \setminus (\tilde{V} \cap V_p)$ back into $\mathcal{T'}$. When connecting a novel node to existing vertices, we loosen the graph building criteria by increasing $D_c$ and decreasing $D_m$, especially to accommodate adding locations around sharp turns. 

%%------------------------------------------------- %
\section{EXPERIMENTAL RESULTS} \label{sec:experiment}
%%------------------------------------------------- %

%-----------------------------------------%
\subsection{Setup} \label{sec:setup}
%-----------------------------------------%

We use the Gibson environment~\cite{xiazamirhe2018gibsonenv} both to generate training datasets and to evaluate navigation performance in simulation. We compare our method against SPTM~\cite{savinov2018semiparametric} and ViNG~\cite{shah2021ving}, which adopt similar navigation pipelines, and can also be fairly assessed after training or fine-tuning on data from each target domain. Moreover, we perform our experiments in realistic cluttered indoor environments. We want to highlight the inherent difficulty arising from navigating in cluttered indoor environments, where the agent is required to continuously avoid colliding with obstacles and navigate through small openings (e.g., doors). We collect 288,000 data points from 10 interactive environments from iGibson~\cite{shenigibson} to initially train our model. In contrast, we collect 500,000 data points each for SPTM and ViNG, as they use a random exploration policy and thus need a larger size dataset to ensure sufficient exploration and visual diversity. The width and height of each RGB-D observation are $96 \times 72$. We use the LoCoBot~\cite{locobot} in both simulated and real-world experiments, and we teleoperate it in each test map to collect trajectories for building the graph.\footnote{Additional implementation details, e.g., visual overlap computation, CNN architecture, hyperparameters, environments, etc., can be found in our project page: 
\href{https://montrealrobotics.ca/ltvn/}{https://montrealrobotics.ca/ltvn/}}

% Also, we believe that outdoor scenarios like the one used in ViNG~\cite{shah2021ving} may not necessarily be more difficult, considering the lack of obstacles in the environment. In fact, we highlight the inherent difficulty arising from navigating in cluttered indoor environments, where the agent is required to continuously avoid colliding with obstacles and navigate through small openings (e.g., doors).

%-----------------------------------------%
\subsection{Evaluation Settings} \label{sec:navigation_settings_setup}
%-----------------------------------------%

We evaluate navigation performance to reflect real-world usage: the agent should be able to navigate between any image pairs from the graph, and should not merely repeat path sequences matching the collected trajectories. We pick 10 goal images from different locations that covers major locations in each map, and generate random test episodes. In simulation, we consider navigation as successful if the position and yaw errors from the goal pose are less than $0.72$m and $0.4$ radians. We consider an episode as a failure if the agent collides for more than $20$ times, and if it requires more than $K$ simulation steps.
For real-world experiments, an episode is deemed successful if the robot's final observation has sufficient visual overlap with the goal image. We consider an episode as a failure if the robot collides with the environment, or if it gets stuck for more than 10 minutes.

During operations, if the agent is unable to localize itself or find a path, we rotate it in-place and take new observations until it recovers. In addition, to ensure fair comparison, instead of training an inverse dynamics model for SPTM, we equip SPTM agent with a pose estimator, and use the same position-based feedback controller as ours and ViNG.

%-----------------------------------------%
\subsection{Navigation Performance in Simulation} \label{sec:sim_results}
%-----------------------------------------%

\begin{figure*}[t]
    \centering
    \includegraphics[width=1.0\textwidth]{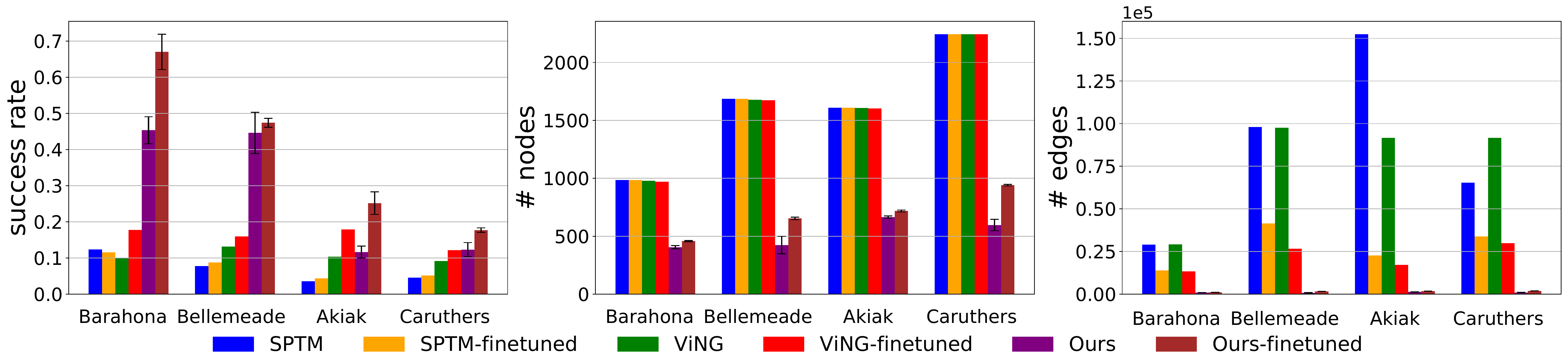}
    \caption{\label{fig:nav_performance} Comparison of navigation success rates and graph sizes among topological visual navigation methods in various test environments. For visual results of our experiments, including real-world deployment videos, see the video attachment or our project page: \href{https://montrealrobotics.ca/ltvn/}{https://montrealrobotics.ca/ltvn/}.}
\end{figure*}

In this section, we compare the navigation performance of our method with SPTM and ViNG in simulated environments both before and after fine-tuning. In addition to navigation performance, we also compare the sparsity of the graphs built with each method. Note that, in this set of experiments, \textit{we do not perform graph maintenance with our method}, which is evaluated separately in Section~\ref{sec:lifelong_results}. 

We evaluate the navigation performance on four unseen test environments: Barahona (57m$^2$), Bellemeade (70m$^2$), Akiak (124m$^2$), and Caruthers (129m$^2$). For trajectory collection, we teleoperate the agent to explore roughly $3-4$ loops around each map, resulting in $985$, $1,685$, $1,609$, and $2,243$ images for Barahona, Bellemeade, Akiak, and Caruthers, respectively. We pick $10$ goal images spanning major locations in each map and generate $500$ random test episodes. Given diverse map sizes, we set $K = 1,000$ for Barahona and Bellemeade, and $K = 2,000$ for Akiak and Caruthers. Since our graph building method is stochastic, we evaluate our method with three random seeds per environment.

As seen in Fig.~\ref{fig:nav_performance}, our method consistently yields higher navigation success rates in all test environments when the model is fine-tuned. We can also observe that the performance gain of our model after fine-tuning is generally higher than others. Additionally, our graphs have significantly fewer number of nodes and edges, which keeps planning costs practical when scaling to large environments. Therefore, compared to the baselines that use entire trajectory datasets to build graphs, our sampling-based graph building method produces demonstrably superior performance and efficiency.

\begin{figure}[h]
    \centering
    \subfigure[SPTM]{\includegraphics[width=0.21\textwidth, angle=90 ,origin=t]{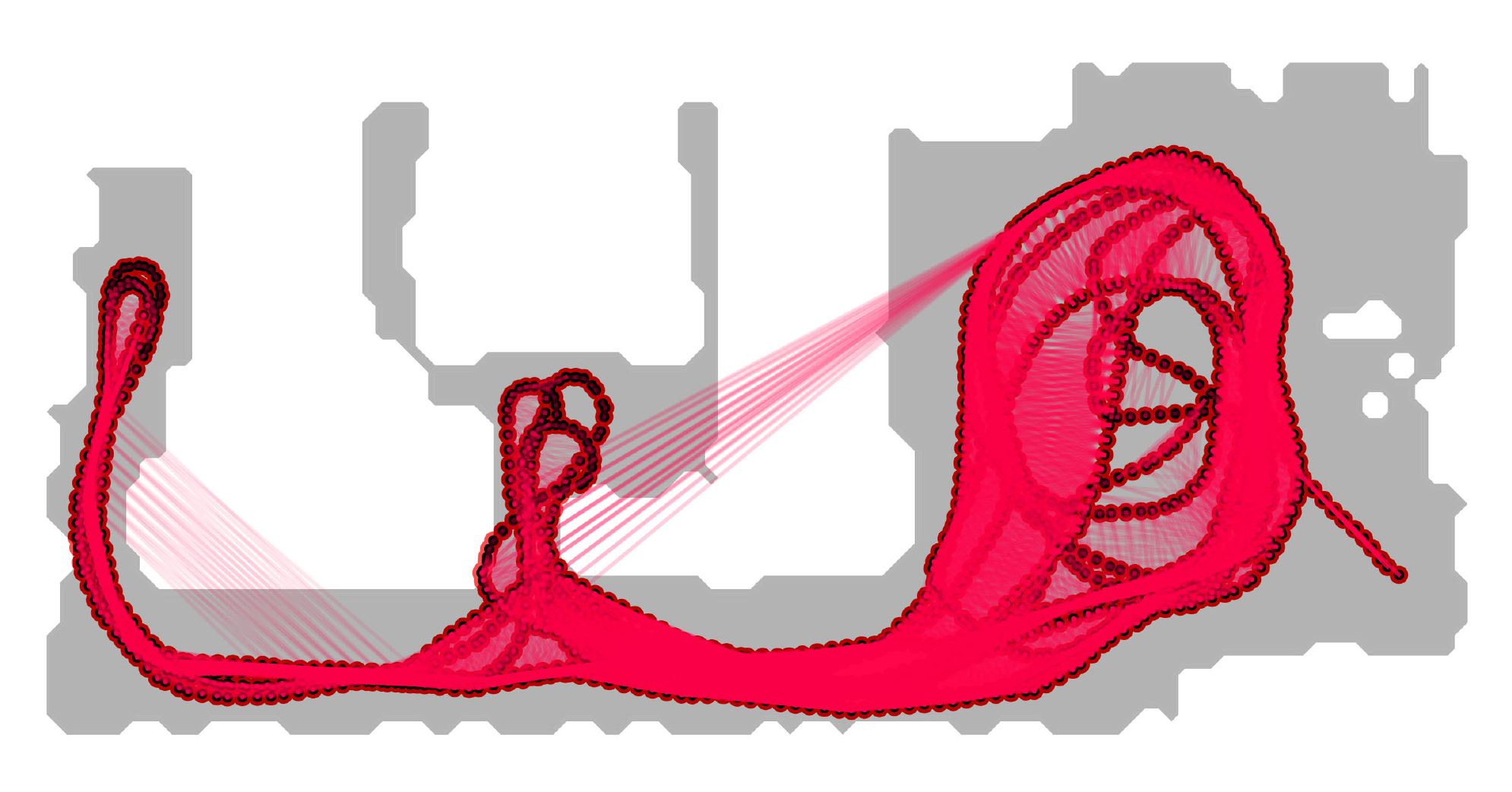}}
    \subfigure[ViNG]{\includegraphics[width=0.21\textwidth, angle=90 ,origin=t]{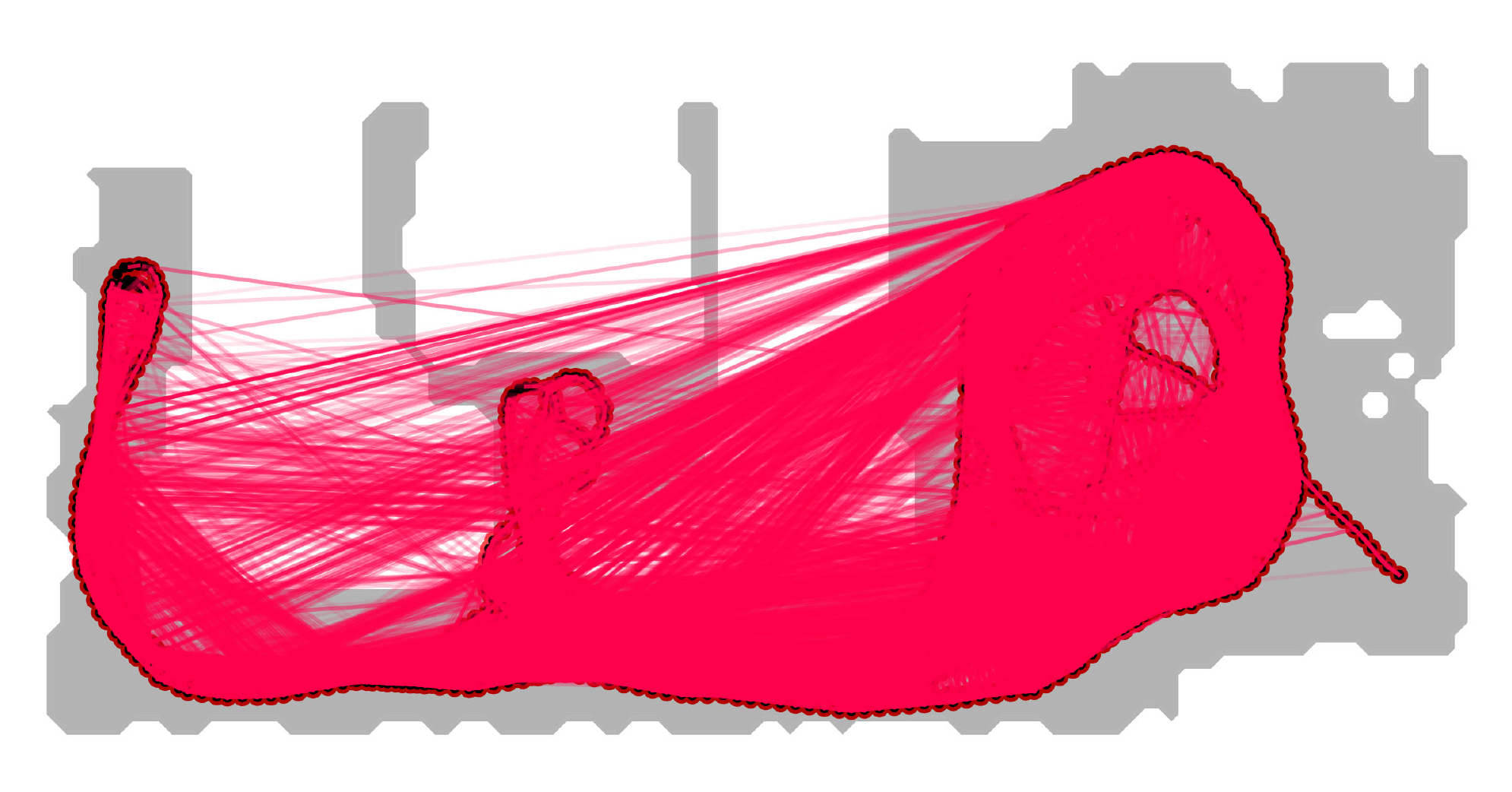}}
    \subfigure[Ours]{\includegraphics[width=0.21\textwidth, angle=90 ,origin=t]{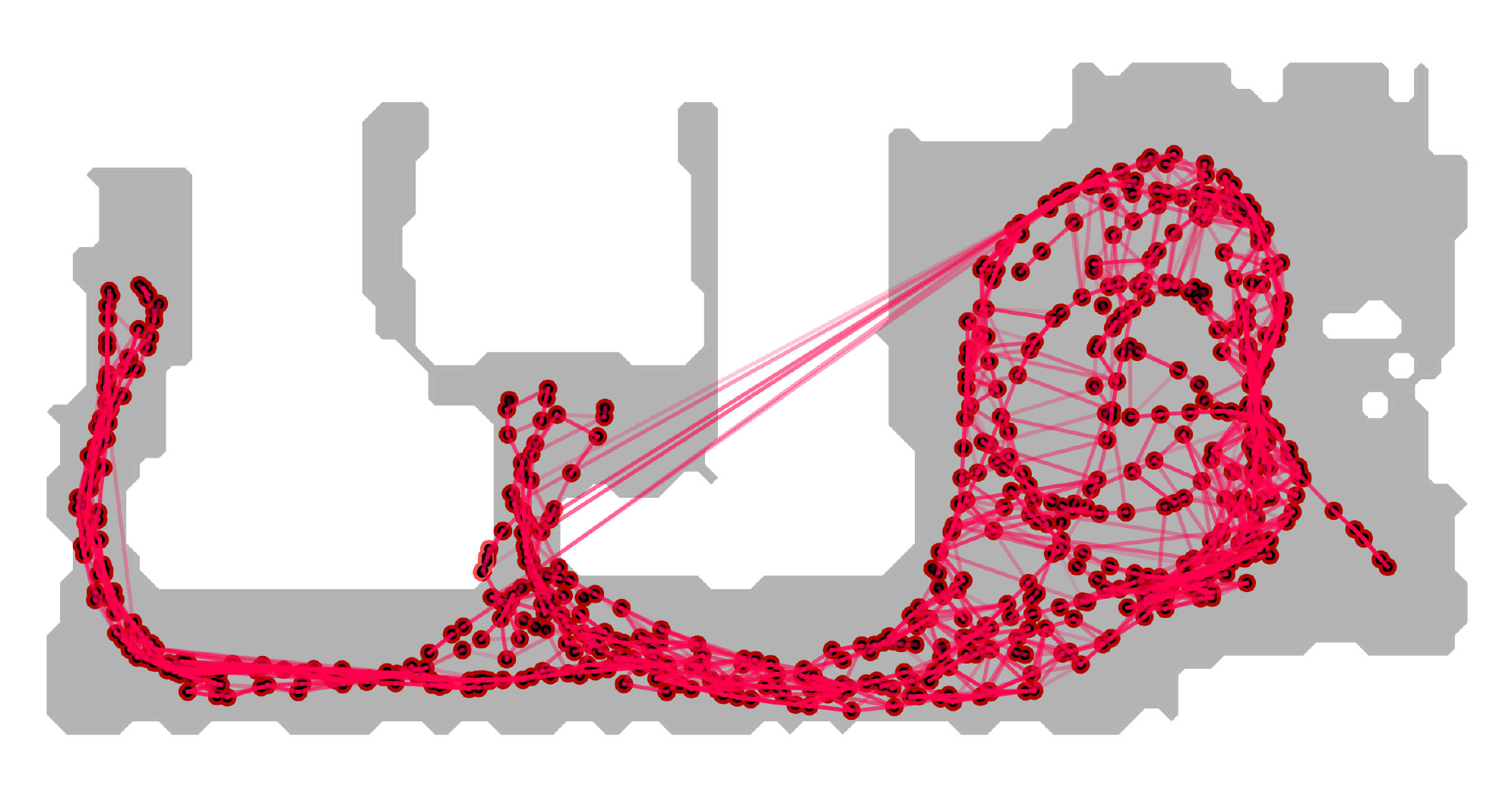}}
    \caption{\label{fig:finetuned_graphs_viz} Graphs built after model fine-tuning in Bellemeade. Even without applying graph maintenance, our method naturally produces a sparser graph.}
\end{figure}

Fig.~\ref{fig:finetuned_graphs_viz} qualitatively compares sample graphs built using different methods. We see that our graph has the fewest vertices, yet still maintains proper map coverage. Visually, our graph also has few false positive edges through walls, and we shall later demonstrate how our graph maintenance can further prune these in Section~\ref{sec:lifelong_results}.

%-----------------------------------------%
\subsection{Lifelong Navigation} \label{sec:lifelong_results}
%-----------------------------------------%

\begin{figure*}[t]
    \centering
    \includegraphics[width=1.0\textwidth]{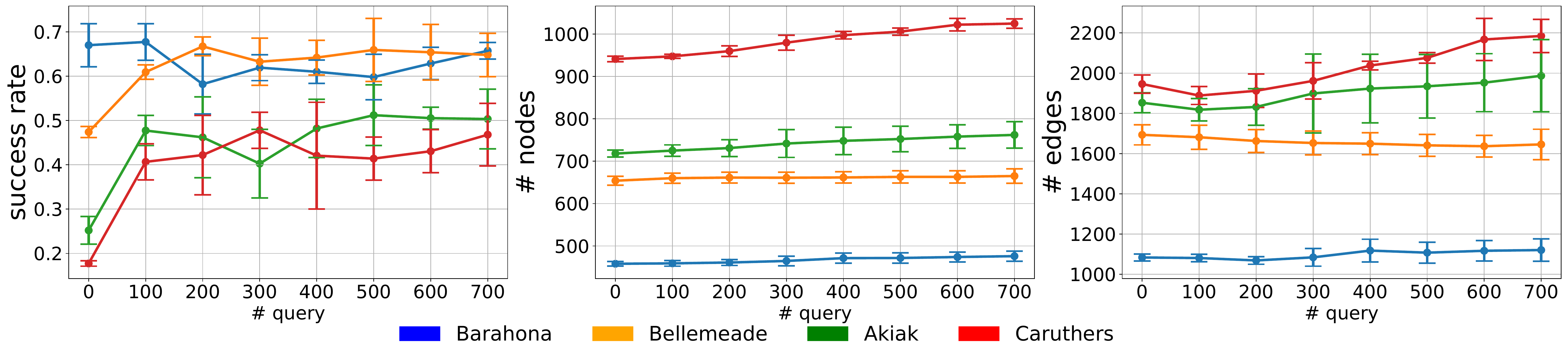}
    \caption{\label{fig:lifelong_results} Changes in success rate, number of nodes, and number of edges as the agent performs more queries and updates its graph in each test environment.}
\end{figure*}

We now evaluate the proposed graph maintenance methods to see how navigation performance is affected as the agent executes more queries. We start these experiments using graphs built with our fine-tuned models. The agent then executes randomly sampled navigation tasks while performing continuous graph maintenance. After every $100$ queries, we re-evaluate the navigation performance on the same static set of test episodes used in Section~\ref{sec:sim_results}. Same as before, we repeat the experiment with three random seeds per environment.

\begin{figure}[h]
    \centering
    \subfigure[No update]{\includegraphics[width=0.21\textwidth, angle=90 ,origin=t]{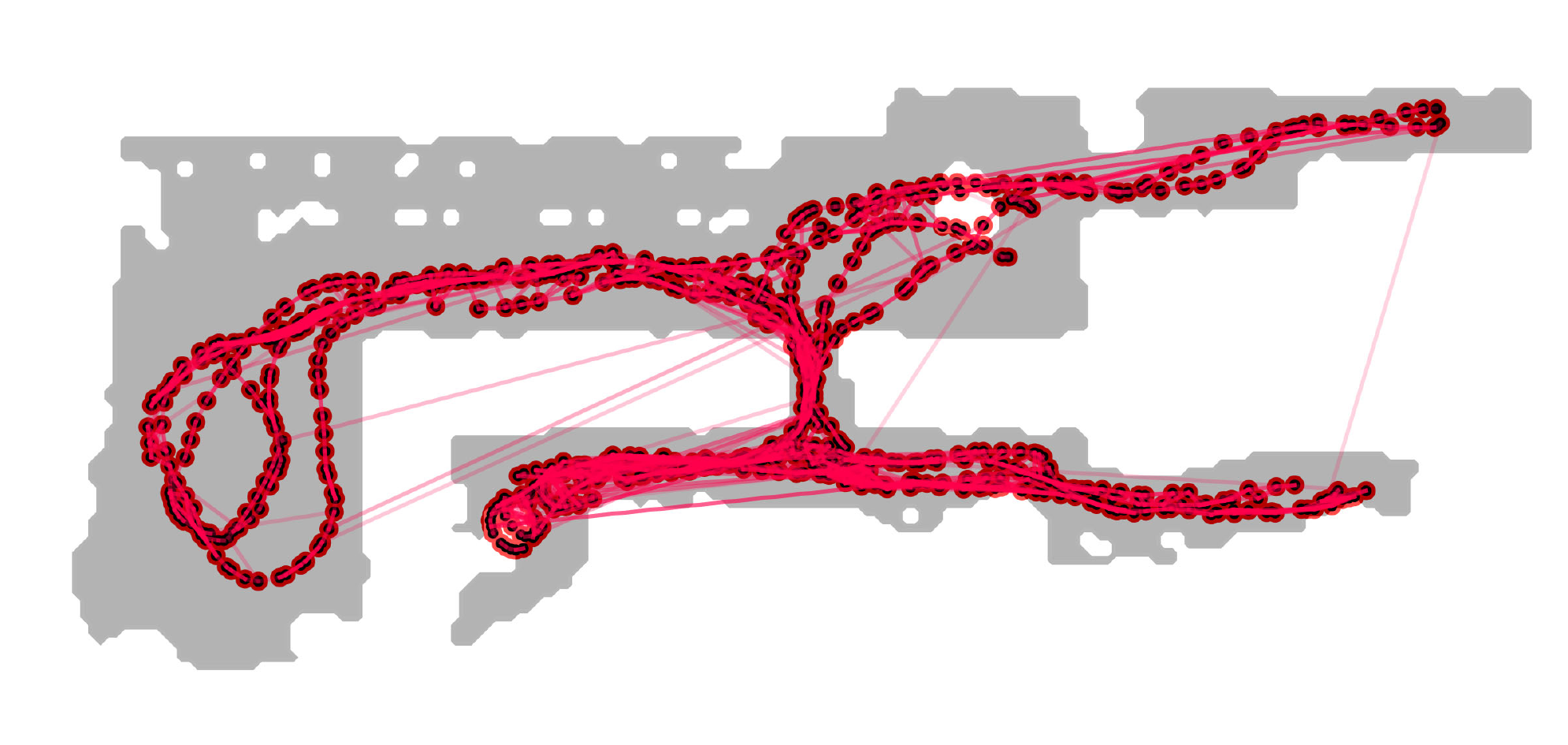}} % \hspace{2cm}
    \subfigure[100 queries]{\includegraphics[width=0.21\textwidth, angle=90 ,origin=t]{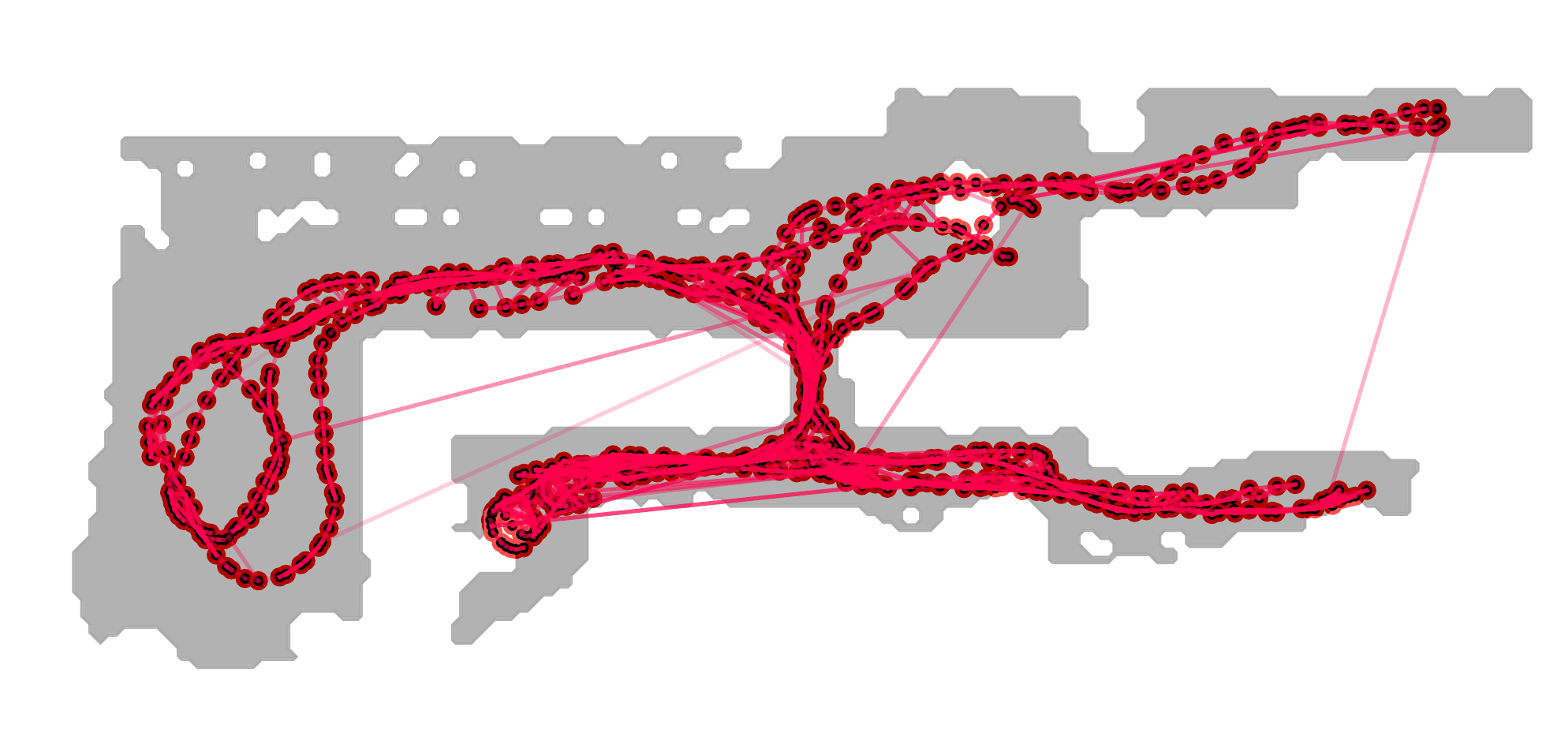}} % \hspace{2cm}
    \subfigure[400 queries]{\includegraphics[width=0.21\textwidth, angle=90 ,origin=t]{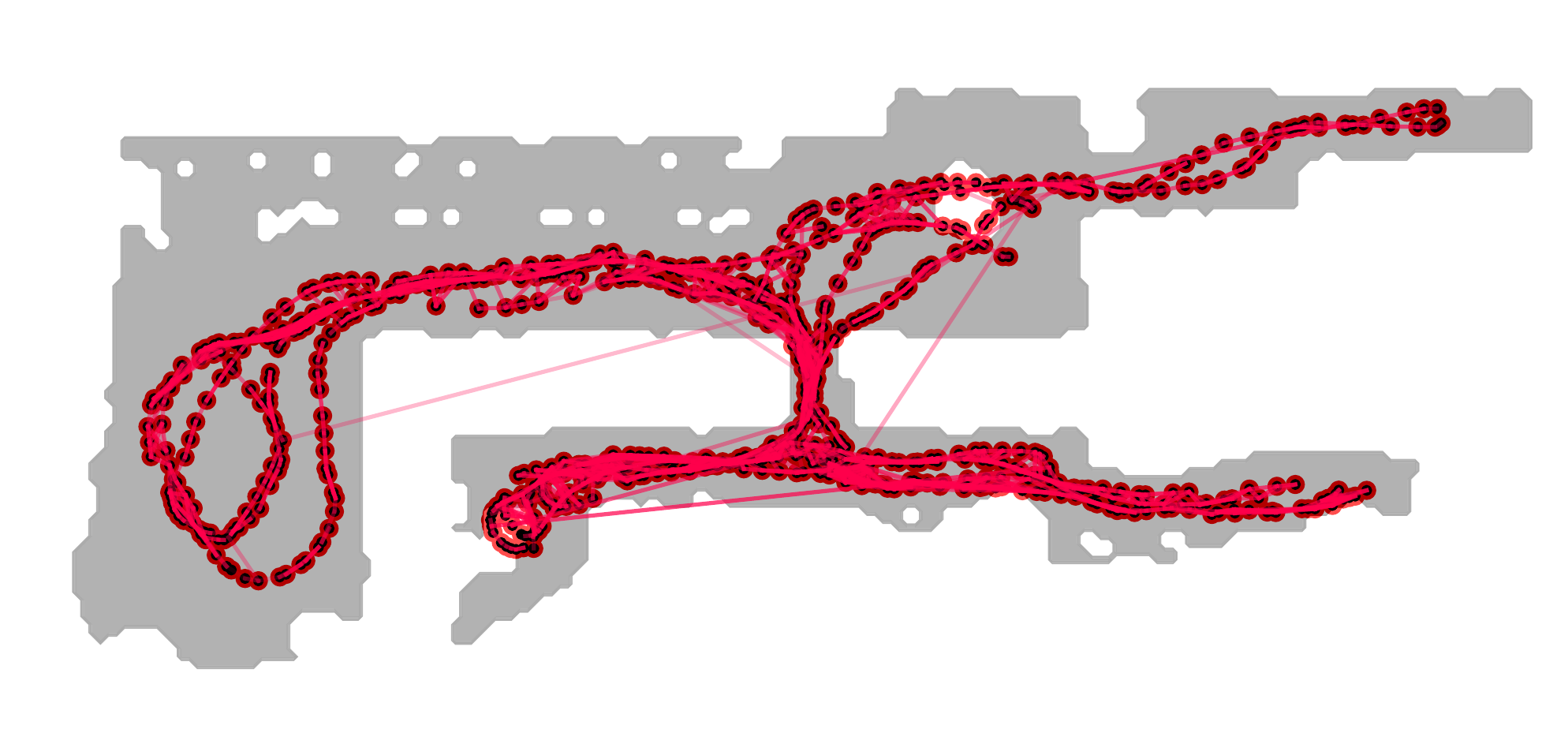}} % \hspace{2cm}
    \subfigure[700 queries]{\includegraphics[width=0.21\textwidth, angle=90 ,origin=t]{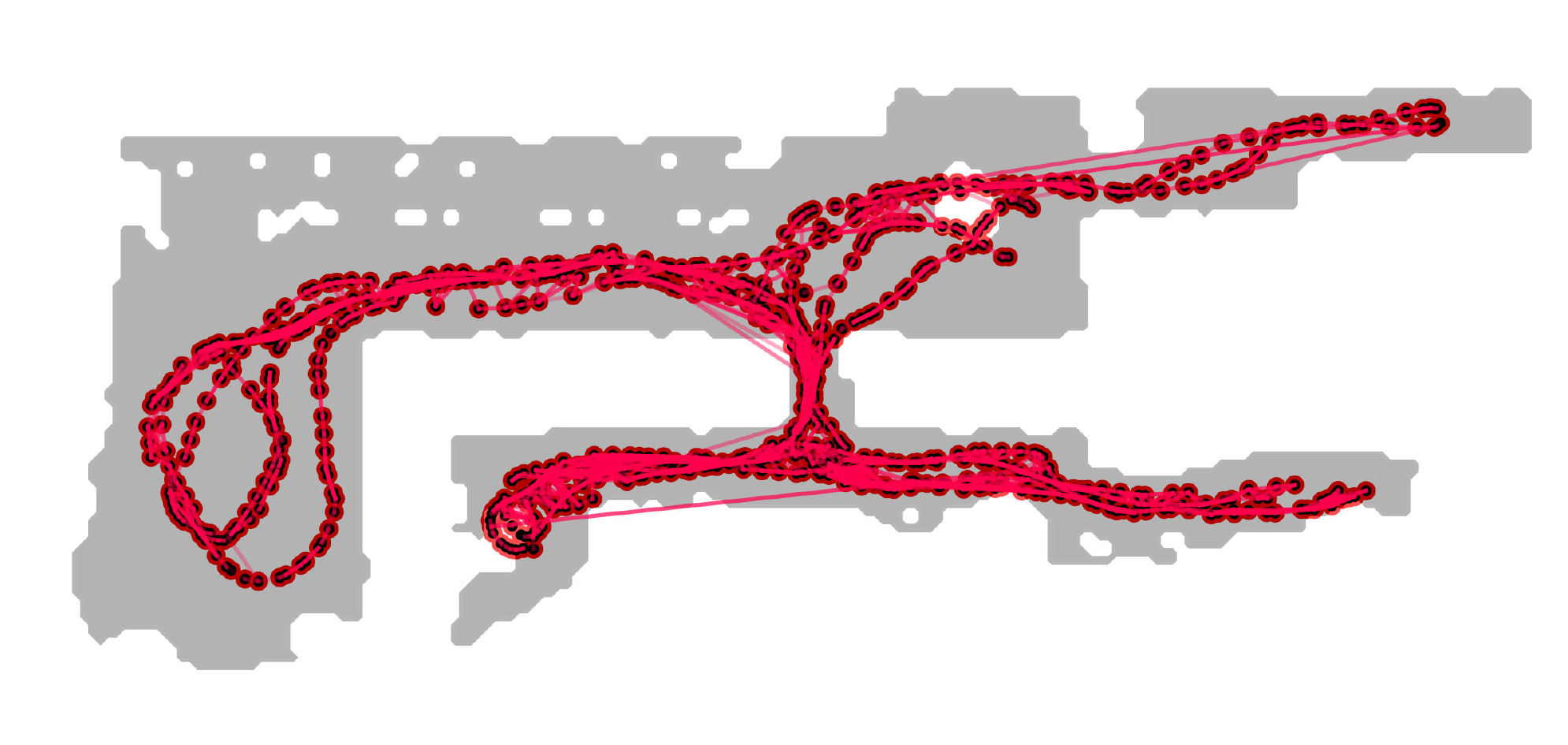}} % \hspace{2cm}
    \caption{\label{fig:updated_graphs_viz} Comparison between the initially built graph and updated graphs after executing 100, 400, and 700 navigation queries in Akiak. We can see a notable reduction in spurious edges, especially ones that are across walls, which improved navigation performance in our experiments.} 
\end{figure}

As seen in Fig.~\ref{fig:lifelong_results}, the success rate initially jumps and continues to generally improve as we perform more queries, while the number of nodes and edges in the graph do not substantially grow. We also see an initial \emph{decrease} in the number of edges, suggesting that our graph maintenance pruned spurious edges causing initial navigation failures, then later added useful new nodes for better navigation. Qualitatively, we can also see fewer spurious edges when comparing sample graphs before and after updates in Fig.~\ref{fig:updated_graphs_viz}.

We observe that sometimes the success rate decreased after a batch of graph maintenance. This is likely caused by new spurious edges when adding new graph nodes near each $100$th query, before we re-evaluate navigation performance. Nevertheless, such spurious edges are pruned in later updates, thus leading to increasing performance trends over time.

%-----------------------------------------%
\subsection{Real-World Experiments} \label{sec:real_world_results}
%-----------------------------------------%

We demonstrate the performance of our method in two real-world environments: a studio apartment and a medium-sized university laboratory. After teleoperating the robot for $3-4$ loops around each space to collect trajectory data, we pick $5$ goal images, and generate $20$ test episodes. We use the iLQR~\cite{li2004ilqr} implementation from the PyRobot~\cite{murali2019pyrobot} library for our controller. In Table~\ref{tab:real_world_results}, we report navigation success rates before and after graph maintenance with $30$ queries.

\begin{table}[h]
    \centering
    \caption{Navigation success rate before and after graph maintenance in real-world environments.}
    \begin{tabular}{|c|c|c|}
        \hline          &          & \\
        [-2.5mm]
          & Before & After \\ %[3mm]
        \hline &  & \\
        [-2.5mm]
        Apartment  & 4/20 & 13/20 \\ %[3mm]
        \hline &  & \\
        [-2.5mm]
        University Laboratory  & 4/20 & 14/20  \\ %[3mm]
        \hline
    \end{tabular}
    \label{tab:real_world_results}
\end{table}

\begin{figure}[h]
    \centering
    \includegraphics[width=0.382\textwidth]{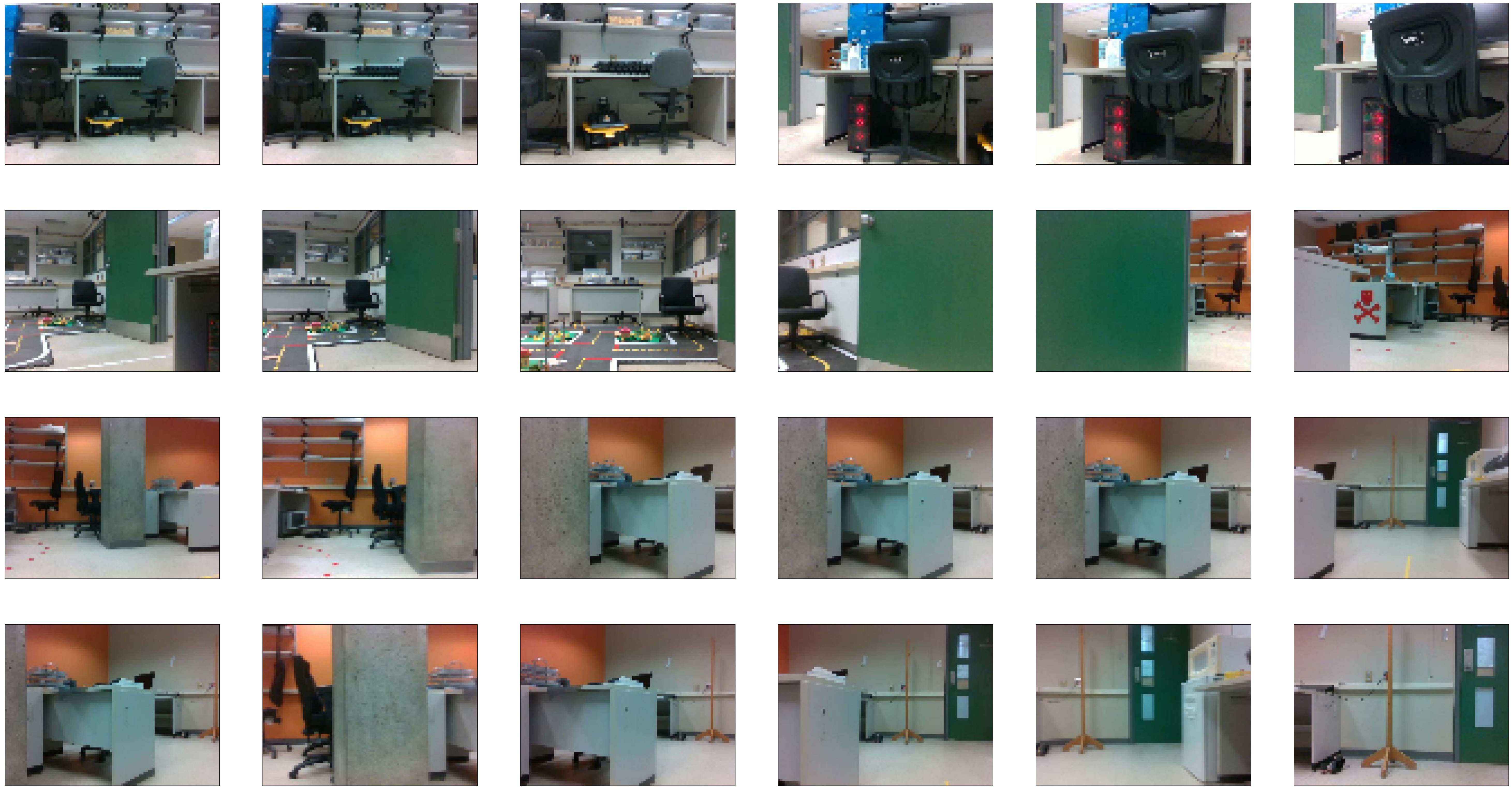}
    \caption{Sequence of navigation images, from top-left to bottom-right, as seen by the robot in the real-world lab environment.}
    \label{fig:real_sample}
\end{figure}

These results suggest that our model performs well without needing large amounts of real-world data, especially when combined with our proposed lifelong graph maintenance. Our graph maintenance enhances the navigation performance with more than $3\times$ increase in success rate in both environments. Fig.~\ref{fig:real_sample} depicts a successful navigation task across multiple twists and turns within the lab environment.

%%------------------------------------------------- %
\section{CONCLUSIONS} \label{sec:conclusion}
%%------------------------------------------------- %

We proposed a simple model that can be used in many topological navigation aspects. With this model, we proposed a new image-based topological graph construction method via sampling, which not only produces sparser graphs compared to baselines, but also higher navigation performance. We also introduced a lifelong graph maintenance approach by updating the graph based on what our agent experienced during navigation. We showed that these updates add useful new nodes and remove spurious edges, thus increasing lifelong navigation performance. We also demonstrated a training regime using purely simulated data, enhanced by fine-tuning on a much smaller dataset from a given target domain, which resulted in strong real-world navigation performance. 

Currently, our model fine-tuning method relies on piloted trajectories with odometry data. It would be more practical if we can fine-tune our model on an unordered set of images, or images taken from different sources such as a mobile phone.
Furthermore, we also assume a static world; extending to non-stationary environments remains a fruitful challenge.

%%------------------------------------------------- %
\section*{Acknowledgments}
%%------------------------------------------------- %

The authors would like to thank Mitacs and Element AI (a ServiceNow company) for the support in this project. R. R. W. thanks IVADO for the support, as well as K. M. Jatavallabhula for useful discussions and feedback. L. P. is supported by the Canada CIFAR AI Chairs Program. The work was also supported by the National Science and Engineering Research Council of Canada under the Discovery Grant Program.

% \section{References Section}
% You can use a bibliography generated by BibTeX as a .bbl file.
%  BibTeX documentation can be easily obtained at:
%  http://mirror.ctan.org/biblio/bibtex/contrib/doc/
%  The IEEEtran BibTeX style support page is:
%  http://www.michaelshell.org/tex/ieeetran/bibtex/
 
% You can manually copy in the resultant .bbl file and set second argument of $\backslash${\tt{begin}} to the number of references
%  (used to reserve space for the reference number labels box).

\bibliographystyle{IEEEtran}
\bibliography{IEEEabrv, ref}

\end{document}